\documentclass[runningheads]{llncs}
\usepackage{graphicx}

\usepackage{times}
\usepackage{amsmath}
\usepackage{soul}
\usepackage{url}
\usepackage[hidelinks]{hyperref}
\usepackage[utf8]{inputenc}
\usepackage[small]{caption}
\usepackage{graphicx}
\usepackage{booktabs}
\usepackage{algorithm}
\usepackage{algorithmic}
\usepackage{amsfonts}
\usepackage{subfigure}
\usepackage[T1]{fontenc}
\usepackage{bm}
\usepackage{verbatim}
\newcommand{\partitle}[1]{\smallskip \noindent \textbf{#1.}}

\usepackage{xcolor}

\begin{document}

\title{Private Semi-supervised Knowledge Transfer for Deep Learning from Noisy Labels}

\author{Qiuchen Zhang\inst{1} \and
Jing Ma\inst{1} \and
Jian Lou\inst{1} \and Li Xiong\inst{1} \and Xiaoqian Jiang\inst{2}}
\authorrunning{Q. Zhang et al.}

\institute{Department of Computer Science, Emory University, Atlanta, GA \\ \email{\{qiuchen.zhang,jing.ma,jian.lou,lxiong\}@emory.edu}\\
\and
University of Texas Health Science Center at Houston, Houston, TX \\
\email{\{xiaoqian.jiang\}@uth.tmc.edu}}

\maketitle              

\begin{abstract}
Deep learning models trained on large-scale data have achieved encouraging performance in many real-world tasks. Meanwhile, publishing those models trained on sensitive datasets, such as medical records, could pose serious privacy concerns. To counter these issues, one of the current state-of-the-art approaches is \textit{Private Aggregation of Teacher Ensembles}, or PATE, which achieved promising results in preserving the utility of the model while providing strong privacy guarantee. PATE combines an ensemble of ``teacher models'' trained on sensitive data and transfers the knowledge to a ``student'' model through the noisy aggregation of teachers' votes for labeling unlabeled public data which student model will be trained on. However, the knowledge or voted labels learned by the student are noisy due to private aggregation.  Learning directly from noisy labels can significantly impact the accuracy of the student model. 

In this paper, we propose the PATE++ mechanism, which combines the current advanced noisy label training mechanisms with the original PATE framework to enhance its accuracy. A novel structure of Generative Adversarial Nets (GANs) is developed in order to integrate them effectively. In addition, we develop a novel noisy label detection mechanism for semi-supervised model training to further improve student model performance when training with noisy labels. We evaluate our method on Fashion-MNIST and SVHN to show the improvements on the original PATE on all measures.

\keywords{Differential Privacy  \and Semi-supervised Transfer Learning \and Noisy Labels}
\end{abstract}

\section{Introduction}
\let\thefootnote\relax\footnotetext{This is the full version of the paper published to \hyperlink{https://ieeexplore.ieee.org/xpl/conhome/9377717/proceeding}{2020 IEEE Big Data}}
Training deep learning models requires large-scale data that may be sensitive and contain user's private information, such as detailed medical histories and personal messages or photographs \cite{miotto2016deep,vinyals2015grammar,parkhi2015deep}. Publishing or sharing those models trained on private data directly could cause information leakage and lead to serious privacy issues, as the adversary could exploit the trained models to infer or reconstruct (the features of) the training data. For example, \textit{membership inference attacks} \cite{shokri2017membership} attempt to infer whether or not a model was trained with a specific record with black-box API access to the model. \textit{Model inversion attacks} \cite{fredrikson2015model} attempt to reconstruct a recognizable face image corresponding to a person (a class label) by analyzing the model parameters of the target face recognition model. 

Differential privacy (DP) \cite{dwork2006calibrating,dwork2014algorithmic} has demonstrated itself as a strong and provable privacy framework for statistical data analysis, and has been utilized to provide strict privacy protection in many practical applications from healthcare data analysis \cite{ma2019privacy} to location-based services \cite{andres2013geo}. It has recently been explored to protect the privacy of training data when training deep learning models \cite{abadi2016deep,papernot2018scalable,10.1093/jamia/ocaa119}. Phan et al. \cite{phan2016differential} explore the objective function perturbation method and use it to train a deep autoencoder satisfying DP. However, it may not be trivial to generalize to other deep learning models. Shokri and Shmatikov \cite{shokri2015privacy} train a deep model across multiple sites collaboratively and protect the DP of each updated parameter. However, the overall privacy budget for the model is huge which leads to a meaningless privacy guarantee. One widely accepted way to provide rigorous DP guarantee for training neural network models on sensitive data is to use differentially private Stochastic Gradient Descent (DP-SGD) which adds Gaussian noise to the gradients in each iteration during the SGD based optimization process \cite{abadi2016deep}. However, as the model goes deeper, this method becomes less effective \cite{papernot2016semi,zhao2019improving}. 

Another promising approach is \textit{Private Aggregation of Teacher Ensembles} (PATE), which trains multiple teacher models on disjoint sensitive data and transfers the knowledge of teacher ensembles to a student model by letting the teachers vote for the label of each record from an unlabeled public dataset \cite{papernot2018scalable}. The teachers' votes are aggregated through a differentially private noisy-max  mechanism, which is to add DP noise to the number of each label's votes first and then take the label with the majority count as the output. Finally, the student model is trained on the partially labeled public dataset in a semi-supervised fashion and published, while the teacher models are kept private.

Compared to DP-SGD, PATE achieves higher accuracy with a tighter privacy guarantee on the same dataset used in both works \cite{papernot2016semi}. Meanwhile, the PATE mechanism is independent of the learning algorithms and can be applied to different model structures and to datasets with various characteristics. However, the knowledge transferred from teachers to the student, which are noisy-max voted labels, contain a certain proportion of errors or noisy labels, and the proportion has a positive relationship with the level of privacy guarantee that PATE provides and a negative impact on the accuracy of the student model. 

To mitigate the dilemma between providing stronger privacy protection and achieving higher model accuracy in the PATE framework and further improve its effectiveness in privacy-preserving model learning, in this paper, we propose an enhanced framework PATE++ by incorporating the state-of-the-art noisy label training mechanism into PATE to further improve its practical applicability. PATE++ makes several novel contributions. First, we modify the student model in the original PATE to enable co-teaching.  The student model of PATE is a generative adversarial network (GAN) \cite{goodfellow2014generative} trained under semi-supervised learning \cite{salimans2016improved}, while co-teaching(+) is originally used for supervised training. To utilize co-teaching(+) in the student model, our main idea is to add an additional discriminator in the GAN to enable co-teaching(+) between the two discriminators for robust training with noisy labels. Second, to further exploit the benefit of semi-supervised training, we propose a novel noisy label detection mechanism based on the co-teaching(+) framework and move the data with detected noisy labels from labeled dataset to unlabeled dataset instead of excluding them completely from the training process.

We evaluate our framework on Fashion-MNIST and SVHN datasets. Empirical results demonstrate that our new PATE++ structure with additional noisy label detection and switching (from labeled data to unlabeled data) mechanism outperforms the original PATE in training deep learning models with differential privacy. Our work further improves the practicality and operability to privately and safely train deep neural network models on sensitive data.


\section{Preliminaries}
In this section, we introduce the definitions of differential privacy \cite{dwork2006calibrating}, and the two essential components of our approach: (1) the PATE framework which was first developed by Papernot et al. in \cite{papernot2016semi} and later improved by Papernot et al. in \cite{papernot2018scalable}; (2) the co-teaching mechanism for robust model training with noisy labels and the improved co-teaching+ mechanism \cite{yu2019does}.

\subsection{Differential Privacy}
Differential Privacy (DP) ensures the output distributions of an algorithm are indistinguishable with a certain probability when the input datasets are differing in only one record, which is achieved through adding some randomness to the output. Both Laplacian noise and Gaussian noise are widely used to achieve DP, and the scale of the noise is calibrated according to the privacy parameter(s) $\epsilon$ (and $\delta$) as well as the sensitivity of the algorithm \cite{dwork2014algorithmic}. 
In the definition of DP, $\epsilon$ and $\delta$ are the privacy parameters or privacy budget, which indicate the privacy loss. A smaller $\epsilon$ means a higher level of indistinguishability and hence stronger privacy. A smaller $\delta$ means a lower probability that the privacy guarantee provided by $\epsilon$ will be broken.

\begin{definition}
(\textit{($\epsilon$, $\delta$)-Differential Privacy}) \cite{dwork2014algorithmic}. Let $\mathcal{D}$ and $\mathcal{D}'$ be two neighboring datasets that differ in at most one entry. A randomized algorithm $\mathcal{A}$ satisfies ($\epsilon$, $\delta$)-differential privacy if for all $\mathcal{S}\subseteq$ Range$(\mathcal{A})$:
$$
Pr\left[\mathcal{A}(\mathcal{D})\in \mathcal{S}\right] \leq e^{\epsilon} Pr\left[\mathcal{A}(\mathcal{D'})\in \mathcal{S}\right]+\delta,
$$
where $\mathcal{A}(\mathcal{D})$ represents the output of $\mathcal{A}$ with the input $\mathcal{D}$. 
\end{definition}

Rényi Differential Privacy (RDP) generalizes ($\epsilon$, 0)-DP in the sense that $\epsilon$-DP is equivalent to ($\infty$, $\epsilon$)-RDP.

\begin{definition}
(Rényi Differential Privacy (RDP)) \cite{mironov2017renyi}. A randomized mechanism $\mathcal{A}$ is said to guarantee ($\lambda$, $\epsilon$)-\textit{RDP} with $\lambda \geq 1$ if for any neighboring datasets $\mathcal{D}$ and $\mathcal{D}'$,
\begin{equation*}
    \begin{split}
        & D_{\lambda}(\mathcal{A}(\mathcal{D}) \| \mathcal{A}(\mathcal{D'})) = \\ &  \frac{1}{\lambda-1} \log \mathbb{E}_{x \sim \mathcal{A}(D)}
\left[\left(\frac{Pr[\mathcal{A}(D)=x]}{Pr\left[\mathcal{A}\left(D^{\prime}\right)=x\right]}\right)^{\lambda-1}\right] \leq \epsilon.
\end{split}
\end{equation*}
\end{definition}

In the above definition, $D_{\lambda}(\mathcal{A}(\mathcal{D}) \| \mathcal{A}(\mathcal{D'}))$ indicates the Rényi divergence of order $\lambda$ between $\mathcal{A}(\mathcal{D})$ and $\mathcal{A}(\mathcal{D'})$.

\begin{theorem} (From RDP to ($\epsilon$, $\delta$)-DP) \cite{mironov2017renyi}. If a mechanism $\mathcal{A}$ guarantees ($\lambda$, $\epsilon$)-RDP, then $\mathcal{A}$ guarantees $\left(\epsilon+\frac{\log 1 / \delta}{\lambda-1}, \delta\right)$-DP for any 0 $<$ $\delta$ $<$ 1.
\end{theorem}

Theorem 1 reveals the relationship between ($\epsilon$, $\delta$)-DP and ($\lambda$, $\epsilon$)-RDP. Both of them are relaxed from pure $\epsilon$-DP, while RDP equipped with Gaussian noise has better composition property when analyzing the accumulated privacy loss.

\subsection{The PATE Framework}

\begin{figure}[ht]
\centering
\includegraphics[width=0.99\linewidth]{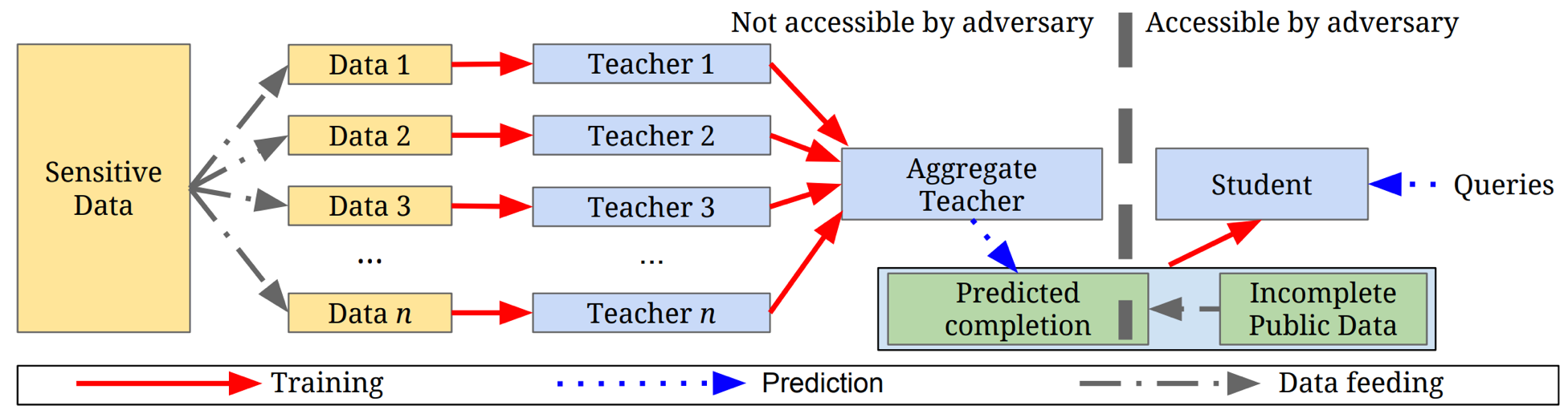}
\caption{Overview of the PATE framework: (1) an ensemble of teachers is trained on disjoint subsets of the sensitive data, (2) a student model is trained on public data labeled using the ensemble.}
\label{fig.pate.framework}
\end{figure}

Figure \ref{fig.pate.framework} illustrates the framework of PATE borrowed from the original paper. It consists of an ensemble of teacher models and a student model. Each teacher is trained on a disjoint subset of sensitive data which contains user's private information that need to be protected. Teacher models can be flexibly chosen to fit the data and task. After teachers are trained, the knowledge that teachers learned from sensitive data will be transferred to the student in a private manner. More specifically, at prediction, teachers independently predict labels for the queried data from an unlabeled public dataset. The votes assigned to each class will be counted to form a histogram. 
To ensure DP, Laplacian or Gaussian noise will be added to each count. The final prediction result for the queried data will be the label with the most votes after adding the noise. 

The student model in the PATE framework uses GAN with semi-supervised learning. During student model training, labeled public data are fed into the discriminator $D$ of GAN to form the supervised cross-entropy loss while unlabeled data and generated data from generator $G$ (labeled as an additional `generated' class) are fed into $D$ to form the unsupervised loss. Feature matching is used to increase the stability of GAN by involving a new objective for $G$, which requires the activations of real data and generated data on an intermediate layer of $D$ to be as similar as possible through gradient-based optimization.

The student model can be published for further use, while the teacher models are kept private. The privacy guarantee provided by the PATE framework can be understood from two aspects. On the one hand, the student model is not trained by sensitive data but through transfer learning. On the other hand, during the knowledge transfer, teachers' votes are perturbed with noise to ensure DP. The initial PATE uses Laplacian noise for the perturbation and moments accountant \cite{abadi2016deep} to compose the total privacy cost for multiple predictions. The improved PATE uses Gaussian noise based on RDP. Additionally, they proposed a \textit{selective} aggregation mechanism called the confident Gaussian NoisyMax aggregator (Confident-GNMax). Teacher ensembles will only answer the queries if their votes have strong consensus, which is checked privately. This mechanism benefits both privacy and utility. The privacy cost is small when most teachers agree on one vote. Meanwhile, when most teachers agree, the prediction result is more likely to be correct. However, even with the Confident-GNMax mechanism, the voted labels still contain a certain ratio of errors due to the noisy aggregation. Additionally, in order to achieve a tighter privacy guarantee, larger noise is needed for perturbing the votes, thus causing more noise in the student training dataset, which severely affects the utility of the trained student model.

\subsection{Co-teaching and Co-teaching+ Mechanisms}
Deep learning models have enough capacity to remember all training instances even with noisy labels, which leads to bad generalization ability~\cite{zhang2016understanding}.
Han et al. \cite{han2018co} propose a simple but effective mechanism called co-teaching for training deep models with the existence of noisy labels. Their method is based on the observation that during the training, models would first memorize or fit training data with clean labels and then those with noisy labels~\cite{arpit2017closer}. Co-teaching maintains two networks with the same structure but independent initialization. In each mini-batch of data, each network selects a ratio of small-loss instances as useful knowledge and teaches its peer network with such useful instances for updating the parameters. Intuitively, small-loss instances are more likely to be the ones with correct labels, thus training the network in each mini-batch using only small-loss instances is more robust to noisy labels.

In the early stage of co-teaching, due to independent and random parameter initialization, two networks have different abilities to filter out different types of error using the small-loss trick. However, this divergence between two networks will gradually diminish with the increase of training epochs, which decreases the ability to select clean data and increases the accumulated error. To solve this issue, Yu et al. introduce the ``Update by Disagreement'' strategy to co-teaching and name the improved mechanism co-teaching+~\cite{yu2019does}. Similar to co-teaching, co-teaching+ maintains two networks simultaneously. In each mini-batch of training, two networks feed forward and predict the same batch of data independently first, and then a ratio of small-loss instances will be chosen by each network only from those data with the disagreed predictions between two networks and fed to each other for parameter updating. This disagreement-update step keeps the constant divergence between two networks and promotes the ability of them to select clean data.

\section{Improved Training Mechanism For PATE}

Inspired by co-teaching mechanism and its improved version co-teaching+, we modify the PATE framework to improve the student models' robustness when training with noisy labels provided by teachers.

\subsection{PATE+: Student Model with Co-teaching+}

\begin{figure}[ht]
\centering
\includegraphics[width=0.99\linewidth]{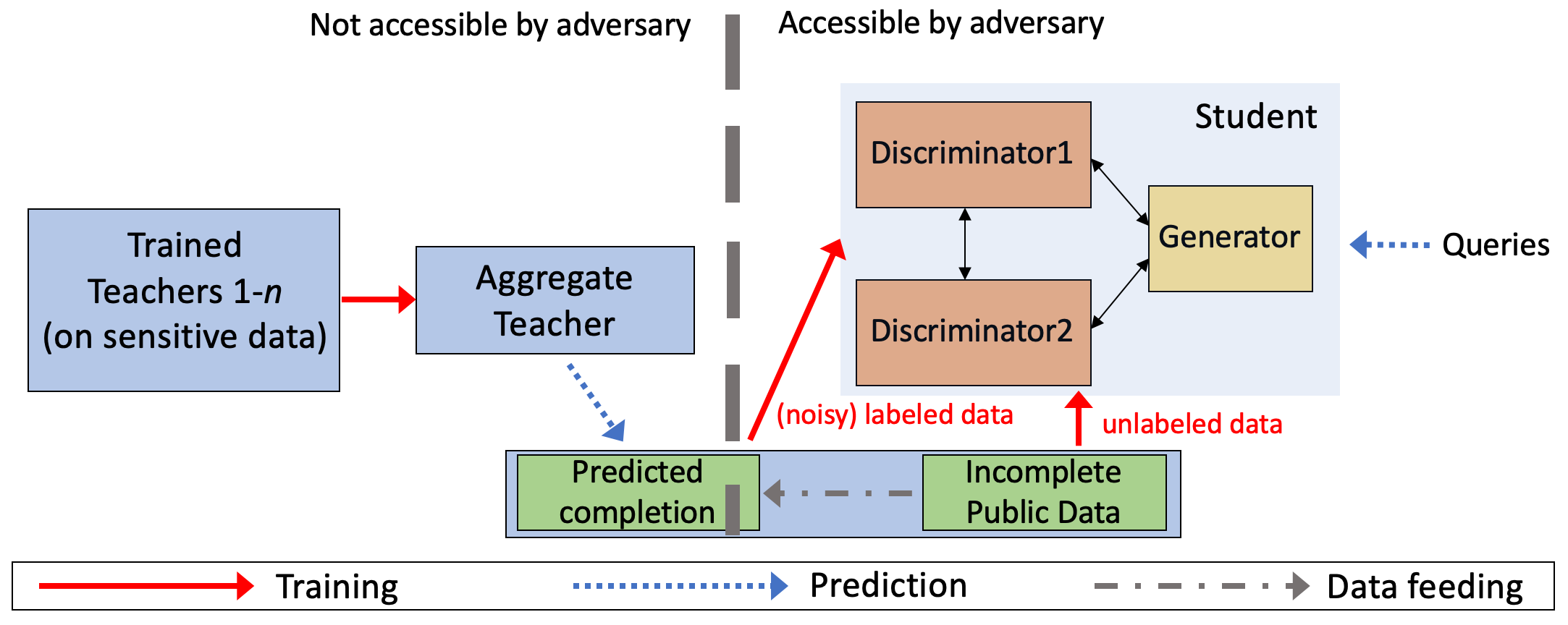}
\caption{Overview of the PATE+ framework. (1) an ensemble of teachers is trained on disjoint subsets of the sensitive data, (2) a semi-supervised GAN student model with one generator and two discriminators co-teaching+ with each other is trained on public data labeled using the ensemble.}
\label{fig.pate++.framework}
\end{figure}

The student model of PATE is a GAN trained under semi-supervised learning with both supervised and unsupervised losses, while co-teaching(+) is originally used in the supervised model training.  To utilize co-teaching(+) in the student model, our main idea is to add an additional discriminator in the GAN used in the student model, as shown in Figure \ref{fig.pate++.framework}. We do not use two GANs with both generator and discriminator as the peers for co-teaching(+) because the small-loss trick plays its role only in the supervised part, while the generator is involved in the unsupervised loss of GAN as well as the feature matching loss \cite{salimans2016improved}, which are both unsupervised and not associated with labels.

Suppose there exist $K$ possible classes in sensitive data as well as the labeled public data that the student model will be trained on. In the semi-supervised learning using GANs, the data generated by generator $G$ are labeled with a new "generated" class $y=K+1$. The discriminator $D$ takes in a data sample $\boldsymbol{x}$ as input and outputs class probabilities distribution $p_{D}(y | \boldsymbol{x}, j<K+1)$. For labeled data $\boldsymbol{x}$, the cross entropy between the observed label and the predicted distribution $p_{D}(y | \boldsymbol{x}, j<K+1)$ forms the supervised loss. For generated data, $p_{D}(y=K+1 | \boldsymbol{x})$ is used to supply the probability that $x$ is not real. For those unlabeled data, since we know they come from one of the K classes of real data, we can learn from them by maximizing $\log p_{D}(y \in\{1, \ldots, K\} | \boldsymbol{x})$~\cite{salimans2016improved}.

For the student model in Figure \ref{fig.pate++.framework}, there are two discriminators and one generator. The supervised loss and unsupervised loss for ${\tt Discriminator1}$ and ${\tt Discriminator2}$ ($D_{1}$ and $D_{2}$) are expressed as:
$$
~~~L_{\text {supervised}}^{D_{i}} ~~= -\{\mathbb{E}_{\boldsymbol{x}, y \sim p_{\text {data }}(\boldsymbol{x}, y)} \log p_{D_{i}}(y | \boldsymbol{x}, y<K+1)\};
$$
\begin{equation*}
    \begin{split}
         L_{\text {unsupervised }}^{D_{i}}=&-\{\mathbb{E}_{\boldsymbol{x} \sim p_{\text {data }}(\boldsymbol{x})} \log \left[1-p_{D_{i}}(y=K+1 | \boldsymbol{x})\right] \\ &~~~~~~+  \mathbb{E}_{\boldsymbol{x} \sim G} \log \left[p_{D_{i}}(y=K+1 | \boldsymbol{x})\right]\}. 
\end{split}
\end{equation*}
where $i=1, 2$ and $p_{\text {data }}$ indicates the real data distribution.

Feature matching loss in the semi-supervised GANs training is defined as:\\ $\left\|\mathbb{E}_{\boldsymbol{x} \sim p_{\text {data }}}  \mathbf{f}(\boldsymbol{x})-\mathbb{E}_{\boldsymbol{z} \sim p_{\boldsymbol{z}}(\boldsymbol{z})} \mathbf{f}(G(\boldsymbol{z}))\right\|_{2}^{2}$, where $p_{\boldsymbol{z}}(\boldsymbol{z})$ indicates the random distribution and $\mathbf{f}(\boldsymbol{x})$ is the activation output of an intermediate layer of the discriminator. In the structure of student model as shown in Figure \ref{fig.pate++.framework}, the generator takes the activations from two discriminators which are expressed as $\mathbf{f}_{D_{1}}(\boldsymbol{x})$ and $\mathbf{f}_{D_{2}}(\boldsymbol{x})$ respectively. We use the average of two feature losses associated with two discriminators as the objective for the generator. Therefore, the feature matching loss of the generator in the student model is defined as:
\begin{equation*}
    \begin{split}
         L_{\text {fm }}^{G} =& \frac{1}{2}  \big{(}\left\|\mathbb{E}_{\boldsymbol{x} \sim p_{\text {data }}} \mathbf{f}_{D_{1}}(\boldsymbol{x})-\mathbb{E}_{\boldsymbol{z} \sim p_{\boldsymbol{z}}(\boldsymbol{z})} \mathbf{f}_{D_{1}}(G(\boldsymbol{z}))\right\|_{2}^{2} \\ &+  
\left\|\mathbb{E}_{\boldsymbol{x} \sim p_{\text {data }}} \mathbf{f}_{D_{2}}(\boldsymbol{x})-\mathbb{E}_{\boldsymbol{z} \sim p_{\boldsymbol{z}}(\boldsymbol{z})} \mathbf{f}_{D_{2}}(G(\boldsymbol{z}))\right\|_{2}^{2}\big{)}.
    \end{split}
\end{equation*}

During the student model updating, a mini-batch of labeled public data is fed forward to $D_{1}$ and $D_{2}$ simultaneously. Each discriminator assigns the predictions and calculates the losses for data in the current mini-batch. A ratio $R$ of data with the smallest losses will be chosen only from those with disagreed predictions by $D_{1}$ and $D_{2}$ respectively and fed to each other for updating parameters. Algorithm \ref{alg:algorithm.pate+} consists of the main steps for training student model where two discriminators conduct co-teaching+ with each other using the ``update by disagreement'' strategy.

\begin{algorithm}[h]
\caption{PATE+: Training Student Model in PATE with Discriminators Co-teaching+}
\label{alg:algorithm.pate+}
\textbf{Input}: $D_{1}$, $D_{2}$, $G$, labeled public data $M_{l}$ from private teachers aggregation, unlabeled data $M_{u}$, batch size $B$, learning rate $\eta$, epoch $E$, ratio $R$. 
\begin{algorithmic}[1] 
\STATE \textbf{Duplicate} $M_{l}$ or $M_{u}$ to make them have the same  size.
\FOR{$e = 1, ..., E$} 
    \STATE \textbf{Shuffle} $M_{l}$, $M_{u}$ into $\frac{|M_{l}|}{B}$ mini-batches respectively.
    
    \FOR{$b = 1, ..., \frac{|M_{l}|}{B}$} 
        \STATE \textbf{Fetch} $b$-th mini-batch $m_{l}$ ($m_{u}$) from $M_{l}$ ($M_{u}$);
        \STATE \textbf{Generate} B fake samples $m_{g}$ from $G$;
        \STATE \textbf{Select} samples with the different predicted results between $D_{1}$ and $D_{2}$ in $m_{l}$ as $\hat{m_{l}}$
        
        \FOR{$i = 1, 2$}
        \STATE \textbf{Fetch} the $R\%$ smallest-loss samples $\hat{m_{l}}^{(i)}$ of $D_{i}$:\\ 
        $
        \hat{m_{l}}^{(i)} = \mathrm{argmin}_{\hat{m_{l}}':\left|\hat{m_{l}}'\right| \geq R |\hat{m_{l}}| } L_{\text {supervised}}^{D_{i}} (\hat{m_{l}}'; D_{i})
        $
        \ENDFOR
        
        
        
        \STATE \textbf{Update} $D_{1} = D_{1} - \eta \nabla (L_{\text {supervised}}^{D_{1}}(\hat{m_{l}}^{(2)}; D_{1}) + L_{\text {unsupervised}}^{D_{1}}(m_{u}, m_{g}; D_{1}))$ // $D_{1}$ indicates parameters of Discriminator1 here 
        \STATE \textbf{Update} $D_{2} = D_{2} - \eta \nabla (L_{\text {supervised}}^{D_{2}}(\hat{m_{l}}^{(1)}; D_{2}) + L_{\text {unsupervised}}^{D_{2}}(m_{u}, m_{g}; D_{2}))$ // $D_{2}$ indicates parameters of Discriminator2 here 
        
        \STATE \textbf{Update} $G = G - \eta \nabla L_{\text {fm }}^{G}(m_{u}, m_{g}; G)$ // $G$ indicates parameters of Generator here
    \ENDFOR
\ENDFOR
\end{algorithmic}
\textbf{Output}: Trained $D_{1}$, $D_{2}$ and $G$, where $D_{1}$ and $D_{2}$ satisfy rigorous DP guarantee.
\end{algorithm}

\subsection{PATE++: PATE+ with Noisy Label Cleansing}
``Update by disagreement'' strategy in co-teaching+ has two potential drawbacks. First, in the late stage of training, two discriminators are going to achieve a similar capacity and consensus on the predictions on most data. Therefore, the number of ``disagreed'' data in each mini-batch is few, which restricts the models from learning since they can only learn from very few data in the late period. Second, the proportion of noisy labels within the ``disagreement'' in the mini-batch is increasing with the epoch, and models' utility is sacrificed by learning from data with more noisy labels. According to the learning pattern of deep models \cite{arpit2017closer}, after the models have learned to fit easy (clean) data, they are more likely to agree on the predictions of clean data while disagreeing on noisy data because the predictions on noisy data have more randomness and errors before models fit them.

\begin{figure}[h]
\centering
\includegraphics[width=0.8\linewidth]{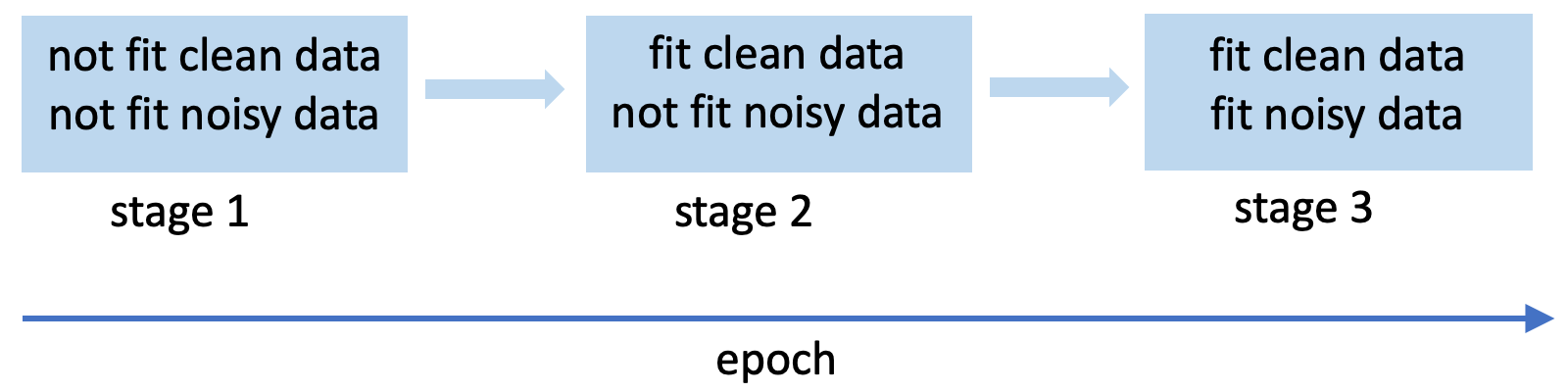}
\caption{Illustration of the three stages of model training process with the existence of noisy data.}
\label{fig.model_training}
\end{figure}

We roughly divide the model learning process into three stages based on the observations in \cite{arpit2017closer}. In the early stage of model training which is indicated as stage 1 in Figure \ref{fig.model_training}, models have not fit either clean or noisy data. Within this stage, the disagreement on predictions between two peer models is mainly caused by the randomness. The percentage of noisy labels within the ``disagreement'' roughly equals the percentage of noisy labels in the whole training dataset. In stage 2, models have fit the clean data (except for ``hard examples'') but not the noisy data. The peer models are more likely to have the same (and correct) predictions for clean data. For those noisy data, since the models have not fitted them, the predictions of them are more random and with more errors. Therefore, prediction disagreements are more likely to happen on the data with noisy labels during this stage. In stage 3, which is the late stage of training, due to the memorization effort, the models have learned to fit both the clean and noisy data. In this stage, the peer models begin to be more consistent in the prediction of both types of data. Thus the ratio of noisy labels in the disagreed predictions is decreased.

Based on this analysis, we hypothesize that the noisy label ratio is the highest within the ``disagreement'' during stage 2.  We propose to filter out noisy labels using this criterion, i.e., the data that has different prediction results by the two peer models during stage 2. However, there is a critical situation that we need to consider. Two peer models do not always have the same learning speed, and they follow different paths during the optimization. Therefore, it could happen that when one model already fits the clean data while the other does not. In this situation, suppose there is a data record with the true (clean) label, the first model gives the correct prediction with high probability, while the second model with the weaker capability predicts it to other labels incorrectly and causes the variation in predictions. Thus clean data could also be chosen by the ``disagreement'' criterion. To exclude this case, we further refine our criterion. Notice that in the above-mentioned circumstance, the ``disagreement'' happens when the first model with the stronger capability predicts the true label for the clean data (the predicted label is the same as the observed label) while the second model with the weaker capability predicts a wrong label (the predicted label is different from the observed label). Therefore, we further filter out noisy data whose predicted labels by peer models are both different from the observed label from the ``disagreement'' in stage 2. That is, our noisy label cleansing mechanism has two criteria: 1) peer models disagree on the predictions for this data, and 2) the prediction results by two peer models are both different from the observed label of the data.

The last question is, how can we know when the models change from stage 1(2) to stage 2(3). One possible method is to use the validation utility to help us decide. In stage 1, models have very low utility since they fit neither clean nor noisy data. In stage 2, the utility of models increases as the models have learned useful knowledge from clean and easy data. In stage 3, models' utility can still increase but with relatively slower speed compared to stage 2, since noisy labels are hard to fit. However, due to the uncertainty of the gradient-based optimization process, it is not efficient to separate these stages using the validation utility. We solve this problem using the weighted decay count. We count the number of epochs for each data when it satisfies the previously mentioned two criteria. Clean label data tend to satisfy those two criteria during stage 1, while noisy data tend to satisfy those two criteria in both stage 1 and stage 2. Therefore, data with more counts at the end of training are determined as the data with noisy labels. To further reduce the effects of stage 1, we multiply a weight (smaller than 1) to the counts at the end of each epoch before adding them to the new counts of the next epoch. Using the weighted decay count smooths the decision process and makes the criteria more robust to the randomness caused by the gradient-based optimization process.

In Algorithm \ref{alg:algorithm.pate++}, we present the complete PATE++ framework for training more robust PATE by filtering out those noisy labels based on the PATE+ framework first, and then retraining PATE+ on the sanitized dataset, which is formed by removing the top  $\tau\%$ data with the most count as introduced above. $\tau$ indicates the removal percentage. We exploit the benefit of semi-supervised learning by moving those filtered noisy data to the unlabeled dataset by removing their labels, since those noisy data themselves have useful pattern information that can be used for the model's training.

\begin{algorithm}[t]
\caption{PATE++: PATE+ with Noisy Label Cleansing}
\label{alg:algorithm.pate++}
\textbf{Input}: $D_{1}$, $D_{2}$, $G$, labeled public data $M_{l}$ from private teachers aggregation, unlabeled data $M_{u}$, batch size $B$, learning rate $\eta$, epoch $E$, ratio $R$, removal percentage $\tau$, decay factor $\alpha$.
\begin{algorithmic}[1] 
\STATE \textbf{Step 1: Filter out noisy label in $M_{l}$ based on PATE+ framework} 
\STATE \textbf{Duplicate} $M_{l}$ or $M_{u}$ to make them have the same  size.
\STATE \textbf{Initialize} the filtered out noisy dataset $M_{n}$ as $\emptyset$.
\STATE \textbf{Initialize} a count table $T$ for each data in $M_{l}$ to be 0.
\FOR{$e = 1, ..., E$} 
    \STATE \textbf{Shuffle} $M_{l}$, $M_{u}$ into $\frac{|M_{l}|}{B}$ mini-batches respectively.
    
    \FOR{$b = 1, ..., \frac{|M_{l}|}{B}$} 
        \STATE \textbf{Fetch} $b$-th mini-batch $m_{l}$ ($m_{u}$) from $M_{l}$ ($M_{u}$);
        \STATE \textbf{Generate} B fake samples $m_{g}$ from $G$;
        \STATE \textbf{Select} samples with the different predicted results between $D_{1}$ and $D_{2}$ in $m_{l}$ as $\hat{m_{l}}$
        \STATE \textbf{Select} samples in $\hat{m_{l}}$ whose prediction results by $D_{1}$ and $D_{2}$ are both different with its observed label as $\overline{m_{l}}$.
        \STATE \textbf{Set} the count of data in $\overline{m_{l}}$ to 1. 
        \STATE \textbf{Fetch} the $R\%$ smallest-loss samples $\hat{m_{l}}^{(1)}$ ($\hat{m_{l}}^{(2)}$) of $D_{1}$ ($D_{2}$) as in line 8-10 in Algorithm \ref{alg:algorithm.pate+}
        \STATE \textbf{Update} $D_{1}, D_{2}$, $G$ as in line 11-13 in Algorithm \ref{alg:algorithm.pate+}
    \ENDFOR
    \STATE \textbf{Multiply} the count of each labeled data in this epoch with $\alpha$ and add to the count table $T$.
\ENDFOR
\newline
\STATE \textbf{Step 2: Remove filtered out noisy labels}
\STATE \textbf{Remove} $\tau\%$ data with the most count from $M_{l}$ to form $M_{l}^{san}$.
\STATE \textbf{Add} those removed data to the unlabeled dataset to form $M_{u}^{san}$.
\newline
\STATE \textbf{Step 3: Retrain the PATE+ on sanitized datasets $M_{l}^{san}$ and $M_{u}^{san}$ using Algorithm \ref{alg:algorithm.pate+}}
\end{algorithmic}
\textbf{Output}: Trained $D_{1}$, $D_{2}$ and $G$, where $D_{1}$ and $D_{2}$ satisfy rigorous DP guarantee.
\end{algorithm}

\section{Experiments}

We performed experiments on Fashion-MNIST and SVHN to demonstrate the efficiency of our proposed PATE+ and PATE++ frameworks compared to the original PATE for training the student model on noisy data provided by private teacher aggregation.

\subsection{Fashion-MNIST}
Fashion-MNIST dataset \cite{xiao2017fashion} consists of 10 classes with 60,000 training examples and 10,000 testing examples. Each example is a gray-scale image with size 28 $\times$ 28. To be consistent with the original PATE paper, we use 60,000 training examples to train the teachers and 10,000 testing examples as the public dataset for training the student. We divide the 60,000 training examples randomly into 250 disjoint subsets equally. Each subset with 240 examples is used to train one teacher model, which is a convolutional neural network with seven convolutional layers followed by two fully connected layers and an output layer (same as the deep model in the original PATE). After 250 teachers are trained, we use Confident-GNMax aggregator to label 2,200 data from the public dataset. 
Adding larger noise during the private teacher aggregation leads to a tighter privacy guarantee (smaller $\epsilon$), while the trade-off is that there will be more noisy labels within the labeled dataset. We study this trade-off and its impact on the student model accuracy.  

The structure of the discriminators in student model is the same as the structure of teachers.  The generator of the student model is a three-layers fully connected neural network. The 10,000 testing examples are further divided into first 9,000 (where 2,200 are labeled by teachers as labeled data and 6,800 are used as unlabeled data) for training and last 1,000 for testing. 
We compare the test accuracy of the student models trained by: 1) the original PATE (traditional semi-supervised training); 2) PATE with co-teaching between two peer discriminators; 3) PATE+ (PATE with co-teaching+ between two peer discriminators); and 4) PATE++ (PATE+ with noisy label cleansing). We train student models on batch of 100 inputs using the Adam optimizer with the learning rate set to 0.01. In PATE++, the decay factor $\alpha$ is set to 0.9 by grid search. 

\begin{figure*}[h]
\centering
\subfigure[$\epsilon$ vs. noisy label rate]{
\includegraphics[width=2in]{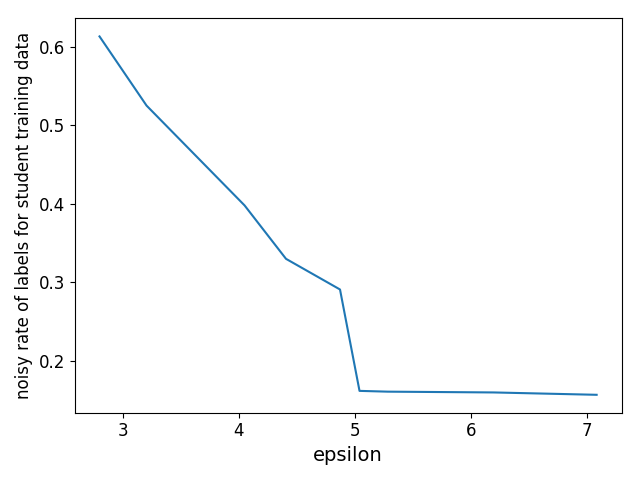}
}
\subfigure[$\epsilon$ vs. test accuracy]{
\includegraphics[width=2in]{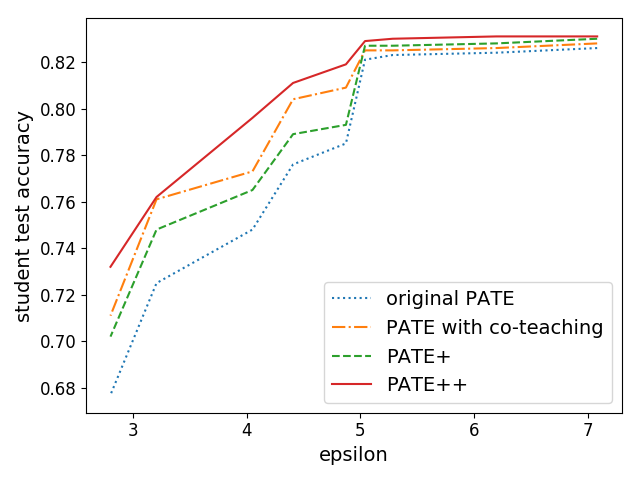}
}
\caption{(a) Relationship between the privacy budget $\epsilon$ (fixed $\delta=10^{-5}$) and the noisy label rate of student training dataset on Fashion-MNIST dataset. (b) Relationship between the privacy budget $\epsilon$ (fixed $\delta=10^{-5}$) and test accuracy of the student model on Fashion-MNIST dataset.}
\label{fig.accuracy.epsilon}
\end{figure*}

Figure \ref{fig.accuracy.epsilon} (a) shows the relationship between the privacy budget $\epsilon$ (fixed $\delta=10^{-5}$) and the noisy label rate of student training dataset from noisy teacher aggregation. The total number of student training data is fixed as 2,200. Different $\epsilon$ values are caused by varying the noise scale of the Gaussian noise when implementing the Gaussian NoisyMax aggregation. The higher $\epsilon$ (the less tight privacy guarantee) indicates the less noise added to the teachers' votes results, therefore, resulting in less noisy student training data. Notice that keeping increasing the $\epsilon$ value will not decrease the noisy label rate to zero because there are intrinsic generalization errors of each individual teacher model as well as their ensemble. Figure \ref{fig.accuracy.epsilon} (b) shows the corresponding student test accuracy within different mechanisms under the varying $\epsilon$ values as in Figure \ref{fig.accuracy.epsilon} (a). We can see that all training mechanisms have higher student test accuracy with higher $\epsilon$ values. Our proposed mechanisms PATE+ and PATE++ outperform the original PATE for all cases.  
The improvement is particularly significant (around 5\%) 
when the privacy budget is tight. 
This further motivates our proposed mechanism PATE++, since there is an inevitable trade-off between utility and privacy in the PATE framework.  The stronger privacy requires larger noise during the private teacher aggregation which leads to higher noisy ratio in the student training data. PATE++ mitigates this by making the student model more robust when trained with noisy labels.


\partitle{Selection of $R$ and $\tau$} As suggested in \cite{han2018co}, the ratio of small-loss instances $R$ should be chosen increasingly during the training since when the number of epochs goes large, the model will gradually overfit on noisy labels. Thus, more instances can be kept in the mini-batch at the start while less should be in the end. We use their proposed scheduling: $R(e) = 1- \beta\min \left\{\frac{e}{15}, 1\right\}$ where $e$ is the epoch and $\beta$ is the estimated noise rate which can be determined by manually verifying a small sampled subset. We report the student accuracy of 1) PATE with co-teaching, and 2) PATE+ (the same as PATE++ with $\tau$=0) under the different estimated noise ratio in Table \ref{tab.varying_R}. We show the setting with (4.05, $10^{-5}$)-DP guarantee. We can see that the estimated noisy rate for the scheduling has effect on the student performance. How to best estimate the noise rate and set the optimal scheduling function is still an unsolved problem in the co-teaching and co-teaching+ works \cite{han2018co,yu2019does}.

\begin{table}
\caption{Test accuracy of the student models with varying $\beta$.}
\label{tab.varying_R}
\centering
\begin{tabular}{|c|c|c|c|c|}
\hline
\textbf{Estimated Noise Ratio} $\boldsymbol{\beta}$ & 0.1 & 0.2 & 0.3 & 0.4 \\
\hline
\textbf{PATE with co-teaching} & 76.2$\%$ & \textbf{77.3$\%$} & 77$\%$ & 76.2$\%$ \\
\hline
\textbf{PATE+ (Alg.\ref{alg:algorithm.pate+})} & 76.4$\%$ & \textbf{76.5$\%$} & 77.3$\%$ & 77.4$\%$ \\
\hline
\end{tabular}
\end{table}

We fix $\beta=0.2$ and report the student accuracy of PATE++ with different $\tau$ values for the noisy label cleansing ratio in Table \ref{tab.varying_tau}. Increasing the removal ratio $\tau$ will increase the chance to move more noisy labels from the labeled dataset to the unlabeled dataset and lead to better student performance because the student model is trained on the dataset with less noisy labels. However, the trade-off is that with the higher removal ratio, less labeled data will be left as well as data with clean labels which student can learn useful knowledge from. In practice, we choose the removal ratio by grid search.


\begin{table}
\caption{Test accuracy of the student models with varying $\tau$}
    \label{tab.varying_tau}
    \centering
    \begin{tabular}{|p{2.8cm}|c|c|c|c|c|c|}
    \hline
    \textbf{Removal Ratio} $\boldsymbol{\tau}$ & 0.091 & 0.182 & 0.227 & 0.273 & 0.318 & 0.364 \\
    \hline
    \textbf{PATE++ ($R=0.2$)} & 78.1$\%$ & 78.6$\%$ & 79$\%$ & 79.5$\%$ & \textbf{79.6$\%$} & 78.2$\%$  \\
    \hline
    \end{tabular}
\end{table}

\subsection{SVHN}
SVHN is a real-world image dataset \cite{netzer2011reading} which has 10 classes with 73,257 training examples and 26,032 testing examples. Each example is a 32$\times$32 RGB image. We use the same structure for the student model as in Fashion-MNIST experiments. The 26,032 testing examples are divided into 10,000 for student training and 16,032 for student testing. We use the clean teacher votes made available online by the authors of PATE to do the Confident-GNMax aggregation for labeling student training data. 3,000 data are labeled privately twice using the smaller noise corresponding to (4.93, $10^{-6}$)-DP guarantee and the larger noise corresponding to (3.96, $10^{-6}$)-DP guarantee respectively. The student models are trained on batches of 100 inputs using the Adam optimizer with the learning rate set to 0.003 and the decay factor $\alpha$ set to 0.9 in PATE++. Table \ref{tab.acc_svhn} shows the experimental results on SVHN where the estimated noise rate $\beta$ is set to 0.2 for scheduling $R$ and the removal percentage $\tau$ is set to 0.4.

\begin{table}
\caption{Test accuracy of the students under different frameworks trained on SVHN dataset.} \label{tab.acc_svhn}
    \centering
    \begin{tabular}{|p{2.6cm}|p{1.4cm}|p{2cm}|p{1.5cm}|p{1.6cm}|}
    \hline
    & \multicolumn{4}{|c|}{\textbf{Student Accuracy}} \\
    \hline
    \textbf{Privacy budget} $~~~~~~~~~~(\epsilon,\delta)$ & \textbf{Original PATE} & \textbf{PATE with co-teaching} & \textbf{PATE+ (Alg.\ref{alg:algorithm.pate+})} & \textbf{PATE++ (Alg.\ref{alg:algorithm.pate++})} \\
    \hline
    ~~~(3.96, $10^{-6}$) & ~~80.5$\%$ & ~~~~~86.1$\%$ & ~79.8$\%$ & ~~\textbf{91.5$\%$} \\
    \hline
    ~~~(4.93, $10^{-6}$) & ~91.7$\%$ & ~~~~~92.8$\%$ & ~~91.6$\%$ & ~~\textbf{93.7$\%$} \\
    \hline
    \end{tabular}
\end{table}


We can observe in Table \ref{tab.acc_svhn} that the PATE++ significantly outperforms the original PATE, especially under the tight privacy budget, with 11\% accuracy improvement. The student accuracy of PATE+ is shy when compared with the PATE with co-teaching. The reason could be the drawbacks of "update by disagreement" strategy that we mentioned previously.

\section{Related Work}
Abadi et al. \cite{abadi2016deep} investigate the DP-SGD method to train neural network models with moments accountant for privacy composition. However, as the model goes deeper, their method becomes less efficient. Papernot et al. \cite{papernot2016semi} propose a protocol called PATE to transfer the knowledge from teacher models trained on the private dataset to the student model in a DP fashion. Their method is independent of the learning algorithms and can be applied to different model structures that are suitable for the tasks. However, there is still a trade-off between privacy and utility when training the student model. Yu et al. \cite{yu2019differentially} utilize zero concentrated DP with refined privacy loss analysis under random resampling and achieve almost the same tight privacy loss estimation with the moments accountant method \cite{abadi2016deep}. Meanwhile, they propose an adaptive privacy budget allocation mechanism to further improve the model utility within the same privacy level.

Following the track of improving the original PATE framework, Wang et al. \cite{wang2019enhance} propose to transfer the knowledge learned from a publicly available non-private dataset to the teachers in order to alleviate the problem that the training data assigned for each individual teacher may be not enough to achieve an ideal performance for some complex datasets and tasks. Sun et al. \cite{sun2019private} exploit knowledge distillation \cite{hinton2015distilling} to further transfer the knowledge from teacher ensembles to the student model privately through the representations from intermediate layers of the teacher models. Meanwhile, they propose to use a weighted ensemble scheme to deal with the situation when the training data for different teachers are biased. Berthelot et al. \cite{berthelot2019mixmatch} develop a new semi-supervised learning algorithm called MixMatch, which achieves state-of-the-art performance in several benchmark datasets by combining several dominant approaches for semi-supervised learning together into a unified framework. They demonstrate that MixMatch improves the performance of PATE with respect to the accuracy-privacy trade-off, which is unsurprising because PATE is a general framework with the student model trained by the semi-supervised learning paradigm in order to reduce the total privacy cost induced by each individual query. Any improved semi-supervised learning algorithm is expected to be used to improve the original PATE framework. Different from these previous works, our work improves PATE from another perspective by incorporating the novel noisy label training and cleansing mechanism under the semi-supervised learning framework to improve the student model accuracy without additional privacy budget allocation.

Many data analysis problems rely on high-quality data. Noise and outliers \cite{xiong2006enhancing}, missing values \cite{10.1007/978-3-030-67658-2_28}, and duplicate data \cite{naumann2010introduction} can seriously affect data quality, in turn, data analysis results. Learning with noisy examples is an old problem with a long research history \cite{quinlan1983learning,angluin1988learning}. Currently, training deep learning models using data with noisy labels has received increasing attention  \cite{han2018co,yu2019does,jiang2017mentornet,malach2017decoupling,li2020dividemix,jiang2019hyperspectral,yao2018deep}. Patrini et al. \cite{patrini2017making} propose to correct the predicted labels by the label transition matrix which is estimated by another network to deal with the noisy labels. Jiang et al. \cite{jiang2017mentornet} develop a technique called MentorNet, which pre-trains an extra network and then uses the extra network for selecting clean samples to guide the training of the main network. Malach and Shalev-Shwartz \cite{malach2017decoupling} propose to train two networks simultaneously, and then update models only using the samples that have different predictions from these two networks. A comprehensive review of all the works within this area is beyond the scope of this paper. We refer interested readers to Frénay and Verleysen \cite{frenay2013classification}, Algan and Ulusoy \cite{algan2019image} for more examples of noisy labels learning methods. 
Our proposed mechanisms incorporate the co-teaching and co-teaching+ methods into the PATE framework to better train the student model with noisy labels and achieve promising results. 

\section{Conclusions}

In this paper, we proposed the PATE+ mechanism for robust training of the student model in PATE. Furthermore, we proposed PATE++ mechanism based on PATE+ which combines co-teaching+ between two discriminators within the structure of GAN and noisy label cleansing. Experimental results demonstrate the advance of our mechanisms compared to the original PATE, especially when the privacy budget is tight. Our proposed mechanisms enhance the utility and privacy trade-off in private model training and further improve the practicality to achieve meaningful privacy guarantees when training deep models on sensitive data. 

Our work opens up a new direction to improve the PATE framework. Investigation of other noisy labels training methods to further enhance the PATE will be an interesting research direction. We also leave applying PATE++ to other applications such as sequence-based models as future work.  
Meanwhile, we hope more work can be done 
to further alleviate the privacy-utility trade-off encountered by many privacy-preserving learning algorithms and make them more practical when dealing with real-world and complex tasks.
\section{Acknowledgments}

This research is supported by National Science Foundation under CNS-1952192 and IIS-1838200, National Institutes of Health (NIH) under UL1TR002378 and R01GM118609, and Air Force Office of Scientific Research (AFOSR) DDDAS Program under FA9550-12-1-0240. XJ is CPRIT Scholar in Cancer Research (RR180012), and he was supported in part by Christopher Sarofim Family Professorship, UT Stars award, UTHealth startup, NIH under award number R01AG066749, U01TR002062, and NSF RAPID 2027790.

\bibliographystyle{splncs04}
\bibliography{bibliography}

\end{document}